\documentclass[10pt, a4paper]{article}
\usepackage{lrec}
\usepackage{multibib}
\newcites{languageresource}{Language Resources}
\usepackage{graphicx}
\usepackage{tabularx}
\usepackage{tikz}
\tikzset{
  basic/.style  = {shape=rectangle, rounded corners,
    draw, align=center,
    top color=white, bottom color=blue!20},
  job/.style   = {basic, bottom color=green!20, },
  vacancy/.style = {basic, bottom color=red!20, text width=2.15cm},
}

\usepackage{array,colortbl,multirow,multicol,booktabs,ctable}
\usepackage{soul}

\usepackage{epstopdf}
\usepackage[utf8]{inputenc}

\usepackage{hyperref}
\usepackage{xstring}

\usepackage{algorithm,algpseudocode}
\usepackage{amsmath}
\usepackage{amsfonts}
\usepackage{amssymb}

\algtext*{EndFunction}

\usepackage{color}

\title{Leveraging the Inherent Hierarchy of Vacancy Titles\\for Automated Job Ontology Expansion}

\name{Jeroen Van Hautte, Vincent Schelstraete, Mika\"el Wornoo}

\address{TechWolf, Belgium \\
         \{jeroen,vincent,mikael\}@techwolf.be\\}

\abstract{
Machine learning plays an ever-bigger part in online recruitment, powering intelligent matchmaking and job recommendations across many of the world's largest job platforms. However, the main text is rarely enough to fully understand a job posting: more often than not, much of the required information is condensed into the job title. Several organised efforts have been made to map job titles onto a hand-made knowledge base as to provide this information, but these only cover around 60\% of online vacancies. We introduce a novel, purely data-driven approach towards the detection of new job titles. Our method is conceptually simple, extremely efficient and competitive with traditional NER-based approaches. Although the standalone application of our method does not outperform a finetuned BERT model, it can be applied as a preprocessing step as well, substantially boosting accuracy across several architectures.\\ \newline  }

\begin{document}

\maketitleabstract

\section{Introduction}
Following the advent of online recruitment, the job market is evolving increasingly towards AI-driven personalised treatment of job seekers~\cite{le2014esco}. This personalisation is typically powered through the combination of machine learning models with extensive knowledge bases, developed both in the private \cite{zhao2015skill,neculoiu2016learning} and public \cite{le2014esco,de2015esco} sector. In this setup, ontologies serve an important function: just like real-life job seekers start with a rough estimate of a given vacancy based on its title, job ontologies provide a similar estimate for thousands of job titles. As vacancies often do not describe the full job contents, but rather provide details on top of the background information contained in this estimate, this allows for a richer and more complete view of the job posting at hand.\\

Many of the taxonomies in use today are curated by hand, as opposed to being data-driven -- this allows for overall high quality and carefully considered structure. However, even with great effort their coverage of the job market is still limited. For example, the ESCO taxonomy \citelanguageresource{ESCO} only covers around 60\% of all job postings available in English, with coverage for other languages often being substantially lower. This disadvantage is typically remedied with machine learning based approaches: an embedding is calculated for any given vacancy title, after which the nearest neighbour among the titles in the knowledge base is selected~\cite{neculoiu2016learning}. While this technique generally works well, it has a crucial weakness: if the job title at hand is conceptually new (or unknown), it can never be mapped onto the knowledge base correctly. As such, any blind spot of the curators can be the direct cause of errors made by the system. With occupations and skills changing faster than ever, such a setup cannot be kept up to date by hand, even with extensive resources.\\

\begin{figure}
    \centering
\begin{tikzpicture}[sibling distance=7em]]
  \node[basic] {}
    child { node[job] {Neurologist} }
    child { node[basic] {Manager}
      child { node[job] {Brand Manager} } 
      child { node[job] {HR Manager}
        child { node[vacancy] {Senior HR Manager} 
          child { node[vacancy] {Senior HR Manager at CompanyX} } 
        }
        child { node[vacancy] {Full-time\\HR Manager\\(London)} }
        child { node[vacancy] {(...) \\HR Manager \\(...) } } }
      child { node[job] {(...) Manager} } 
      }
    child { node[basic] {...} }
;
\end{tikzpicture}
\caption{Job titles (green) and vacancy titles (red) tend to follow an intuitive hierarchy based on lexical inclusion.}
    \label{fig:title_tree}
\end{figure}
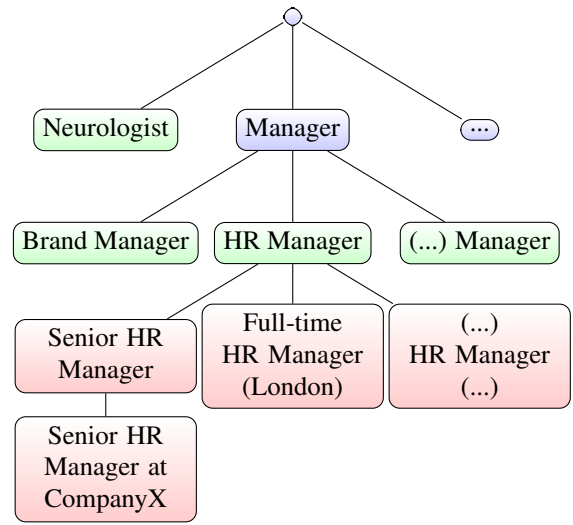

Instead of building knowledge bases by hand, it is also possible to leverage the massive amount of data produced by online recruitment. More precisely, new job titles can be detected from the stream of vacancy titles.\footnote{Throughout this paper, `job title' is used for the name of a function, while a `vacancy title' is the title of a vacancy page -- for example, `digital marketeer' is a job title, while `digital marketeer at Google, London' is a vacancy title.} This problem translates to a typical named entity recognition (NER) setup. While this purely NLP-based approach is often effective, it also largely ignores the underlying structure that holds for job titles. In this paper, we introduce a novel data-driven approach that, using only a large set of vacancy titles, is competitive with conventional neural network-based NER methods. Furthermore, our method can be combined both with these models to gain a substantial performance boost. Our approach is intuitive, lightweight and orders of magnitude faster than competitive models.

\section{Related Work}
\subsection{Job \& Skill Ontologies}
\begin{figure}
    \centering
    \includegraphics[width=0.45\textwidth]{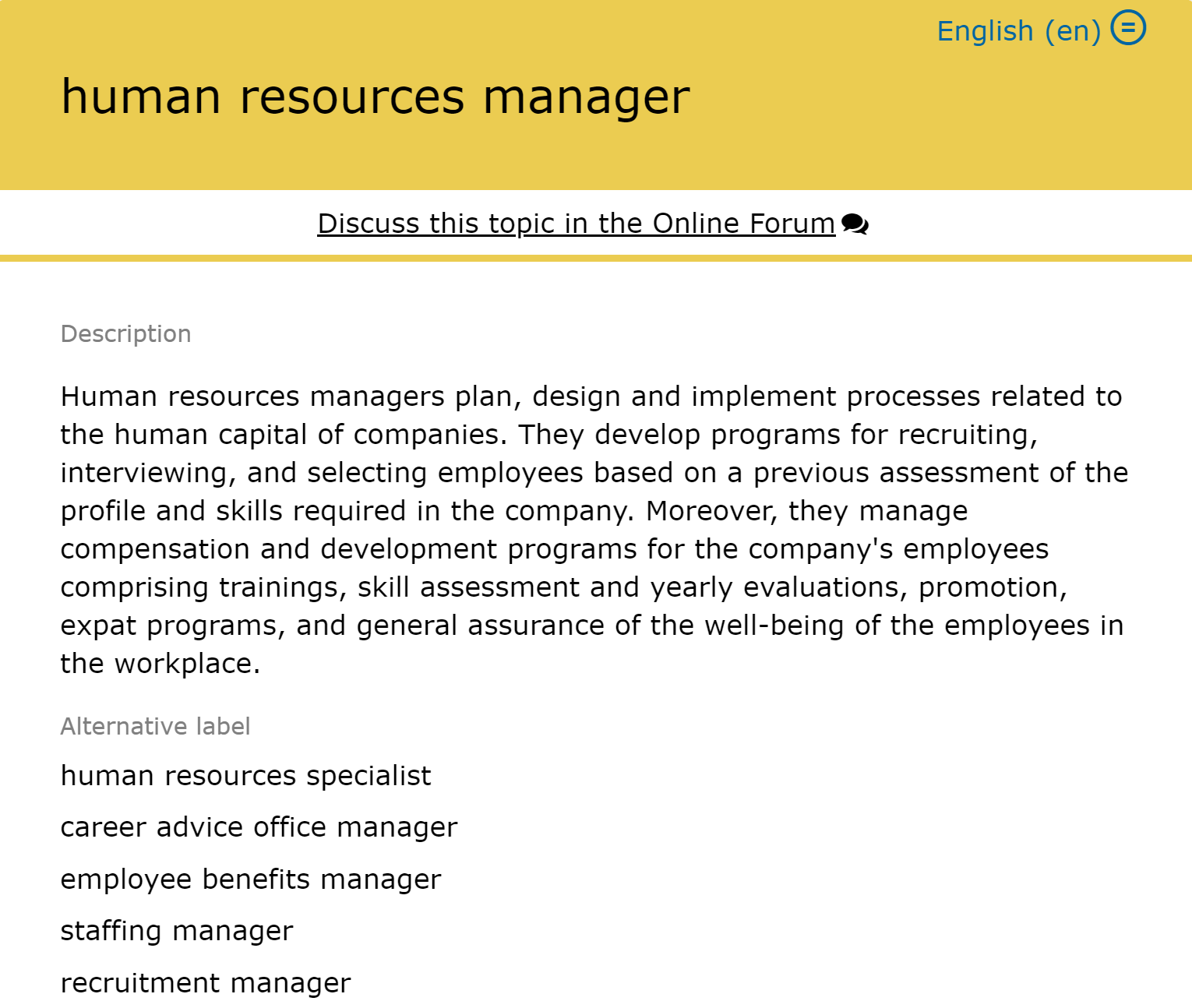}
    \caption{An example of an occupation profile from ESCO. Each occupation has a preferred and alternative labels, a description and a list of optional and essential skills, competences and knowledge.}
    \label{fig:esco}
\end{figure}

The European Skills, Competences, Qualifications and Occupations taxonomy~\citelanguageresource{ESCO} is a handcrafted ontology connecting jobs and skills. It is available in 27 languages and covers close to 3000 distinct occupations, as well as more than 13000 skills. ESCO is funded by the European Commission and is under continuous, active development by its Directorate-General for Employment, Social Affairs and Inclusion. This paper uses version 1.0.3 of the ESCO Classification. Figure \ref{fig:esco} shows an example of an occupation profile -- our setup makes use of the preferred label and alternative labels for each occupation.\\

While ESCO seeks to model occupations and competences at a European level, there are also many alternatives. Each of these has a similar underlying idea, but a different scope or execution strategy. For example, the United States has its O*NET classification \cite{peterson2001understanding}, while France has the ROME standard and the Flemish employment agency VDAB has its own, ROME-based competency standard Competent. Although the experts composing these ontologies leverage data to compose their standards, none of them is data-driven: instead, occupation profiles are typically determined per sector by relevant experts.

\subsection{NER Models}
We compare and combine our novel method with two Named Entity Recognition models: an Iterated Dilated Convolutional Neural Network (ID-CNN) \cite{Strubell_2017} as implemented in SpaCy \cite{honnibal2017spacy} and a fine-tuned BERT model \cite{devlin2019bert} based on the popular transformers library \cite{wolf2019transformers}. In both cases, we make use of an IOB named entity tagging scheme.

\subsection{Automatic Term Recognition}
Finding new job titles in a stream of vacancy titles is a form of automatic term recognition. However, typically this field focuses on finding terminology inside long, grammatical documents rather than titles. \newcite{Frantzi:ECDL1998} use statistical properties of domain-specific language to detect terms in a corpus using their C-value technique. An important principle leveraged in their work is the occurrence of nested terms: terms tend to occur in other, longer terms (as a substring). A useful term is then characterised by its `independence' from longer terms: if something can be used as a term independently, it typically occurs in a larger number of different longer phrases. Since its publication, the C-value/NC-value technique has been applied broadly for detection of multiword expressions, as well as ontology population and expansion based on free text \cite{petasis2011ontology}. Lexical inclusion relations have also been found to account for a substantial part of hierarchical relations among medical concepts \cite{Grabar:COMPUTERM2002}, showing that these principles can be leveraged to construct an accurate hierarchy at a relatively low computational cost.

\subsection{Job Title Detection \& Classification}
Detecting new job titles and assigning job titles to existing classes are two closely related problems. However, as ontologies have largely been composed manually, the focus of most relevant research has been on the latter: instead of using machine learning to build a structure, the techniques are leveraged to position new samples inside the existing hierarchy. For example, \newcite{javed2016towards} use a hierarchical classification system to link job titles to the O*NET classification, using the Lingo algorithm \cite{osinski2005concept} to generate a title hierarchy, after which the formed clusters are assigned to different O*NET concepts. Building upon this work, \newcite{wang2019deepcarotene} use a single end-to-end multistream CNN architecture to classify titles, leveraging both vacancy titles and descriptions. \newcite{neculoiu2016learning}, using a different approach, train a siamese neural network to specifically embed vacancy titles in such a way that relevant job title information is prioritised. This network is then used to map titles onto a proprietary ontology. As related work is generally closed-source, only has a high-level description or does not include an evaluation dataset, we are unable to compare our work with it directly.\\

\section{Method}
\subsection{Job Titles}
For this inquiry, we define a job title for a vacancy to be the minimal subspan of the vacancy title that is needed to determine to which occupation inside ESCO it can be linked. For example, for a vacancy titled ``Senior HR Manager at CompanyX'', the job title would be ``HR Manager''. Modifiers to the job title that concern seniority, practical details or other information are not needed to classify a job within ESCO, as opposed to the words selected. We assume that a job title is always a single, connected span.

\subsection{Title Trees}
\label{section:title_trees}
An important assumption in treating the problem of labelling vacancies with job titles as a NER problem is that inside each vacancy title, a correct job title is present as a subspan. In practice, a vacancy title might not contain a job title (or could contain multiple), but this assumption holds for an overwhelming majority of online job postings, with exceptions typically being poorly composed titles. For example, many of these nonconforming titles are made up of a single, often nonsensical word, most likely provided as a way to fill in a required field, rather than with the intent of informing job seekers. Looking beyond these exceptions, we find a simple, yet interesting hierarchy among job and vacancy titles, as shown in Figure~\ref{fig:title_tree}. In this structure, the parent-child relationship is that of lexical inclusion: a parent is always a substring of each of its children.\footnote{For simplicity, the figure shows a single parent for each title -- in practice, multiple copies of the same title can exist for different parents.} As we move deeper into the tree from the root node, the titles encountered grow increasingly specific, as the addition of more information to a title narrows its scope. Following such a path, there are three types of nodes encountered, following a set order:
\begin{enumerate}
    \item \textbf{Pre-title nodes}: these nodes are parts of job or vacancy titles, but are not valid titles themselves. For example, ``Manager'' or ``Junior'' are part of this category.
    \item \textbf{Job title nodes}: these nodes are both valid job and vacancy titles. Some cases, such as ``Neurologist'', have no parents other than the root node, while others, such as ``HR Manager'', do.
    \item \textbf{Vacancy title nodes}: these nodes are valid vacancy titles, but not valid job titles. They are almost\footnote{Looking at large numbers of online vacancies, we observe that job titles that are frequent enough always occur as standalone vacancy titles.} always inside a subtree that has a job title node at its root.
\end{enumerate}

Given a set of unlabelled vacancy titles, we can construct this tree structure easily by checking which titles contain which other titles. The problem of finding a job title within a given vacancy title is then reduced to finding the right ancestor for this vacancy title (or possibly the title itself). The tree can be implemented efficiently as a trie. In this structure, each node is represented by an ordered sequence of words, with the root being the empty sequence. To insert a new title starting at a given node, its sequence is compared to that of each child. If a child sequence is contained in the current title, the process is continued starting from this child. When no such child can be found, the title is added as a new child to this node. The construction of this trie has a complexity of $M log(N)$, where $M$ is the maximal number of words per title and $N$ is the number of unique titles inside the data structure. By inserting the titles in the order of their number of tokens, each title can be inserted as a leaf node, reducing the implementation complexity substantially.

\subsection{Title Occurrence Ratio (TOR)}
With this title tree, we have now created a setup very similar to the one used by \newcite{Frantzi:ECDL1998} for their C-Value/NC-Value method. However, while the latter uses a collection of n-grams generated from a longer text, this situation involves a large number of much shorter documents. This exposes an essential incompatibility of the C-value method with vacancy titles: while the C-Value is very suitable to distinguish between pre-title and job/vacancy title nodes, the difference between the latter two is much harder to assess, as both job titles and long vacancy titles get very high C-Values. Using a minimum count and maximum length can provide some relief but does not remove the problem entirely. Using the same principles as \newcite{Frantzi:ECDL1998}, we therefore introduce the Title Occurrence Ratio (TOR), which reflects the ratio between how often a title occurs as a standalone vacancy title, and how often it occurs in general (including appearances as a substring of a vacancy title). Unlike the C-Value method, our approach does not treat stop words or certain part-of-speech tags differently, as this was found to make no difference for our use case. The \textit{GetRatio} function in the algorithm below shows how to calculate the ratio for a given title, leveraging the trie data structure described in the previous subsection. Note that for efficiency, the different calls to \textit{BuildTrie} can be replaced by a single, pre-built trie structure.\\

\begin{algorithmic}[1]
\Require{$T_{1} \dots T_{N}$ (normalised vacancy titles)} 
\Require{$Counts$ (a dictionary with the count for each title)}
\Require{$VacTitle$ (the vacancy title at hand)}
\Ensure{$JobTitle$ (the predicted job title subspan)}
\Statex
\Function{GetParents}{$Title, T[\;]$}
  \State $Trie \gets BuildTrie(T[\;]) \text{ // Build a trie with all titles.}$
  \State $Anc \gets Trie.extract(Title) \text{ // Find all ancestors.}$ 
  \State $PN \gets \{\} \text{ // Initialise parent nodes as empty.}$ 
    \For{$X$ in $sort(Anc, key \text{=} \lambda X \xrightarrow{} - X.length)$}
        \State $PN.add(X)$
        \State $PN \gets PN - GetParents(X, Trie)$
    \EndFor
    \State \Return {$PN$}
\EndFunction\\

\Function{GetRatio}{$Title, T[\;]$}
  \State {$C_0 \gets Counts[Title]$}
  \State {$C_1 \gets 0$}
    \For{$X$ in $GetParents(Title, T[\;])$}
        \State $C_1 \gets C_1 + Counts[X]$
    \EndFor  
    \State \Return {$\frac{C_0}{C_0+C_1}$}
\EndFunction\\

\Function{GetJobTitle}{$VacTitle, T[\;]$}
  \State $Trie \gets BuildTrie(T[\;])$
  \State $Cand \gets Trie.extract(VacTitle) + \{VacTitle\}$ 
  \State {$Cand.filter(\lambda X \xrightarrow{} R_{min} < GetRatio(X) < R_{max})$}
  \State \Return {$max(Cand, \lambda X \xrightarrow{} GetRatio(X))$}
\EndFunction

\end{algorithmic}

\subsection{The TOR Method}
\begin{figure}
    \centering
    \includegraphics[width=0.5\textwidth]{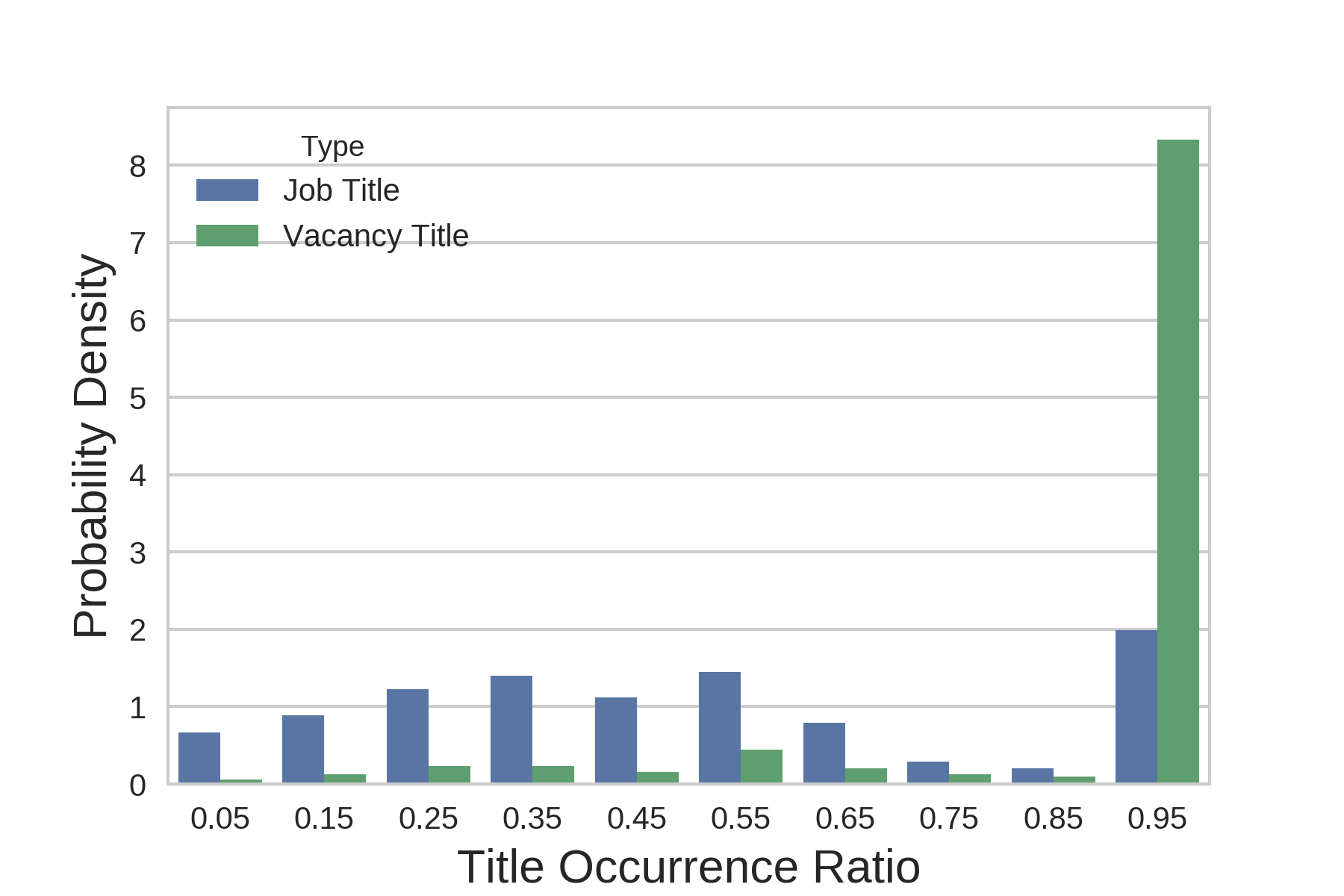}
    \caption{The probability distribution of both job and vacancy titles over their Title Occurrence Ratio.}
    \label{fig:tor_prob}
\end{figure}

We now propose our novel job title extraction method based on this ratio. As figure \ref{fig:tor_prob} shows, the general distribution of vacancy title ratios (in green) differs greatly from that of job titles (in blue). While it is not possible to separate the two based on this number alone, vacancy titles tend to have a ratio close to one, while job titles have a much softer distribution centred around 0.45. It should be noted that the vacancy title distribution contains a component that looks much like the job title distribution -- this is potentially linked to job titles not included in the ESCO dataset. Similarly, there are job titles with a very high TOR, which are most likely to be rare job titles that do not occur more than a handful of times within our dataset.\\

As described in Section \ref{section:title_trees}, a path from root to leaf can be seen as having up to three phases, with the job title phase (which we want to select) lodged in the middle. As the title ratio typically increases steadily from root to leaf, we aim to build a very simple selection system by placing an upper and lower bound on the ratio. Both of these boundaries are optimised using a labelled training dataset, after the construction of the title tree using the combined training and test set. With these selection boundaries in place, the job title for a given vacancy title is now predicted to be its closest ancestor that does not violate the upper and lower bound. Our method is applied as a standalone technique, as well as to preprocess titles before feeding them to the CNN and BERT models.

\begin{table*}[ht]
\small
    \centering
    \setlength\tabcolsep{3pt} 
    \begin{tabular}{l|cccc|cccc}
    \toprule
    & \multicolumn{4}{c|}{\textbf{Micro Average}} & \multicolumn{4}{c}{\textbf{Macro Average}}\\
    \textbf{Method} & Precision & Recall & $F_1$ & Title Acc. & Precision & Recall & $F_1$ & Title Acc.\\
    \midrule
    Identity Baseline & 0.33 & \textbf{1.00}* & 0.20 & 0.02 & 0.53 & \textbf{1.00}* & 0.70 & 0.25 \\
    CValue & 0.78 & 0.90 & 0.83 & 0.59 & 0.77 & 0.56 & 0.65 & 0.30\\
    CNN & 0.89 & 0.82 & 0.85 & 0.67 & 0.89 & 0.79 & 0.84 & 0.61 \\
    BERT & 0.93 & 0.94 & 0.93 & 0.81 & 0.94 & 0.89 & 0.92 & 0.71 \\
    \midrule
    $\text{TOR}_{1M}$ & 0.88 & 0.91 & 0.90 & 0.72 & 0.81 & 0.50 & 0.62 & 0.18 \\
    $\text{TOR}_{100M}$ & 0.85 & 0.93 & 0.89 & 0.68 & 0.86 & 0.79 & 0.82 & 0.59 \\
    $\text{TOR}_{1M}$ + CNN & 0.85 & 0.93 & 0.89 & 0.73 & 0.88 & 0.84 & 0.86 & 0.64 \\
    $\text{TOR}_{1M}$ + BERT & \textbf{0.94} & 0.95 & \textbf{0.94} & \textbf{0.84} & \textbf{0.95} & 0.90 & \textbf{0.93} & \textbf{0.74} \\
    \bottomrule
    \end{tabular}
    \caption{Evaluation results on the constructed task -- the best result in each column is marked in bold. (*) Recall of the identity baseline is 1 by construction.}
    \label{tab:results}
\end{table*}

\section{Evaluation}
The goal of our system is to find new or unknown job titles within a stream of vacancy titles. We measure the success of each approach by evaluating how well it manages to extract job titles from their respective vacancy titles. We make use of two separate types of metrics:
\begin{itemize}
    \item \textbf{Title level metrics}: the main metric is the \textbf{title level accuracy}, which measures how often a fully correct title for a vacancy was extracted. This is the most direct representative for the actual value of a system in practice, as high accuracy is required to be able to contribute to an ontology.
    \item \textbf{Token level metrics}: while the title level accuracy allows for the best performance ranking, insights on the token-level predictions for each method can prove valuable as well. By measuring how well each system predicts whether a token in the vacancy title is part of the corresponding job title, we can gain a better understanding of its behaviour. For example, a system might have low title level accuracy due to a bias towards longer titles, which can be easily read from the token level precision and recall.
\end{itemize}

For each metric, we calculate both the micro and macro average (grouped by the job title label), as to be able to compare performance for frequent and rare job titles. Our main metric, title level prediction accuracy, corresponds directly to a large part of the value of our system in a practical context, as it is only possible to gain useful information about new and unknown titles if they are extracted from vacancies correctly. As to mimic this scenario for our evaluation setup, we separate ESCO into a training set (the set of known titles) and a test set (the set of new/unknown titles). We make sure to avoid these sets influencing each other directly, by ensuring there are no lexical inclusion relations between members of different sets. Using a sample of 1 million scraped vacancy titles\footnote{From company websites and job boards in the UK.}, we now select the vacancies containing each of these titles, using the contained job title as the gold standard.\footnote{While this annotation can cause errors in some cases, it resolves the problem of collecting sufficient annotated data.} We find that in 57.4\% of all vacancies, an ESCO title is included in the title -- vacancies where no match could be found are kept separately in the background set. While this background set is not a part of the training or test set, we include it for the training phase of the TOR method, as to make sure that the evaluation task does not have a bias towards methods based on lexical inclusion properties. In our final dataset, the training set contains 124~108 unique vacancy titles, while the test set contains 45~647 vacancy titles.\\

We evaluate two separate versions of the TOR method: $\text{TOR}_{1M}$, which is trained on the original set of 1 million vacancy titles (including the training, test and background set) and $\text{TOR}_{100M}$, which is trained on a much larger set of 100 million vacancy titles. Optimising on the training set, we find optimal ratio boundaries of 0.03 and 0.69. $\text{TOR}_{100M}$ is only applied as a standalone model, to reflect performance changes when more data is added. For the NER methods, only the longest continuous span of tokens marked as a job title by the model is used as a prediction, as a fragmented prediction would always be counted as an error due to the construction of our dataset. We also include two baselines: the identity baseline, which predicts the entire vacancy title to be part of the job title, and the C-Value method by \newcite{Frantzi:ECDL1998}, using an optimal minimum count of 5 and C-Value threshold of 0.

\section{Results}
The results for the job title extraction task are shown in Table \ref{tab:results}. Consistent with earlier work~\cite{devlin2019bert}, the BERT model substantially outperforms the CNN both in terms of micro and macro average. While the C-Value method outperforms the identity baseline, it generally lags behind other methods across the board. Our novel TOR method is competitive with the neural methods, with both $\text{TOR}_{1M}$ and $\text{TOR}_{100M}$ outperforming the CNN in terms of micro-average. $\text{TOR}_{1M}$ exhibits a clear performance decrease for rare titles, as shown by its low macro averaged scores. However, feeding the same algorithm with 100 million vacancy titles instead, scores show a substantial boost. The TOR method is over 100 times faster than both BERT and the convolutional model, as well as having a smaller memory footprint. This makes our method especially interesting for applications with strict timing requirements or massive amounts of data. For applications where timing is of lesser importance, the TOR method can still be beneficial: the hybrid models, combining TOR with a more typical NER model, show consistent performance improvements across the board. This is especially clear in the improved title-level accuracy, showing that the inherent hierarchical structure of job and vacancy titles can be leveraged to improve general-purpose models. Our method is extremely efficient, compatible with any NER method and easy to implement, making for an easy way to improve job matching systems. By construction, the evaluation setup reflects the discovery of previously fully unknown job titles, showing that these methods are of particular interest for the (semi-)automated expansion of job market ontologies, leveraging data-driven insights to keep standards up to date in a job market that is changing faster than ever. During the review phase for this paper, we applied our method at the behest of VDAB, the Flemish employment agency. In this project, our technique was used to suggest new titles for its Competent standard. As Competent is written in Dutch, we used the RobBERT model introduced by \newcite{delobelle2020robbert}. We found results to be comparable to those obtained in English on the ESCO ontology, with the main difference being a higher macro averaged score, likely to be the consequence of the different methodology used to construct Competent. These results show that our method generalises across multiple languages and occupational taxonomies.

\section{Conclusion}
While the current trend of ever-bigger NLP models does result in the promised performance gains, we have shown that a simple technique incorporating domain knowledge can provide a further boost to the task of extracting job titles from vacancy titles. Our method is conceptually simple, over two orders of magnitude faster than competing models and can be applied in tandem with more general NER models. While our technique struggles with rare job titles when trained on a small dataset, this issue disappears when more data is added, with the TOR method achieving performance comparable to a CNN. Aside from using our method as a standalone model, it can also be leveraged as a preprocessing step, consistently resulting in improved accuracy. Future work will explore the application of our method in different fields, as well as more advanced ways to leverage the title tree used in this paper.

\section*{Acknowledgements}
We thank the anonymous reviewers for their valuable feedback. This publication uses the ESCO classification of the European Commission. The application of the techniques described in this paper to the Competent standard is part of a project funded by VDAB.

\section*{Bibliographical References}\label{reference}

\bibliographystyle{lrec}
\bibliography{lrec2020W-xample-kc}

\section*{Language Resource References}
\label{lr:ref}
\bibliographystylelanguageresource{lrec}
\bibliographylanguageresource{languageresource}

\end{document}